\documentclass[runningheads,a4paper]{llncs}

\usepackage{amssymb}
\usepackage{graphicx}
\usepackage{multirow}
\usepackage{float}
\usepackage[outerbars]{changebar}
\usepackage{amssymb,amsmath}
\usepackage{dsfont}

\usepackage{listings}
\usepackage{url}
\usepackage[pagebackref=false,pdfborder={0, 0, 0},pdftex, bookmarks=2]{hyperref}
\hypersetup{
    pdftitle={Datum-Wise Classification: A Sequential Approach to Sparsity},
    colorlinks,%
    citecolor=black,%
    filecolor=black,%
    linkcolor=black,%
    urlcolor=black
}
\DeclareMathOperator*{\argmax}{argmax}
\DeclareMathOperator*{\argmin}{argmin}

\begin{document}

\mainmatter 

\title{Datum-Wise Classification:\\A Sequential Approach to Sparsity}
\titlerunning{Datum-Wise Sparse Classification}

\author{Gabriel Dulac-Arnold\thanks{This work was partially supported by the French National
Agency of Research (Lampada  ANR-09-EMER-007).}\inst{1} \and Ludovic Denoyer\inst{1} \and Philippe Preux\inst{2} \and Patrick Gallinari\inst{1} }
\institute{
Universit\'e Pierre et Marie Curie - UPMC, LIP6\\
Case 169 - 4 Place Jussieu - 75005 Paris, France\\
\email{firstname.lastname@lip6.fr}
\and
LIFL (UMR CNRS) \& INRIA Lille Nord-Europe\\
Universit\'e de Lille - Villeneuve d'Ascq, France\\
\email{philippe.preux@inria.fr}
}
\index{Dulac-Arnold, Gabriel}

\maketitle
\begin{abstract}
We propose a novel classification technique whose aim is to select an appropriate representation for each datapoint, in contrast to the usual approach of selecting a representation encompassing the whole dataset.  This \textit{datum-wise} representation is found by using a sparsity inducing empirical risk, which is a relaxation of the standard $L_0$ regularized risk.  The classification problem is modeled as a sequential decision process that sequentially chooses, for each datapoint, which features to use before classifying.  Datum-Wise Classification extends naturally to multi-class tasks, and we describe a specific case where our inference has equivalent complexity to a traditional linear classifier, while still using a variable number of features.   We compare our classifier to classical $L_1$ regularized linear models ($L_1$-SVM and LARS) on a set of common binary and multi-class datasets and show that for an equal average number of features used we can get improved performance using our method.
\end{abstract}

\section{Introduction}
\vspace{-0.30cm}

Feature Selection is one of the main contemporary problems in Machine Learning and has been approached from many directions. One modern approach to feature selection in linear models consists in minimizing an $L_0$ regularized empirical risk.  This particular risk encourages the model to have a good balance between a low classification error and high sparsity (where only a few features are used for classification). As the $L_0$ regularized problem is combinatorial, many approaches such as the LASSO \cite{Tibshirani1994} try to address the combinatorial problem by using more practical norms such as $L_1$. These approaches have been developed with two main goals in mind: restricting the number of features for improving classification speed, and limiting the used features to the most useful to prevent overfitting.  These classical approaches to sparsity aim at finding a sparse representation of the features space that is global to the entire dataset.


We propose a new approach to sparsity where the goal is to limit the number of features \textbf{per datapoint}, thus \textit{datum-wise sparse classification\/} (DWSC).
This means that our approach allows the choice of features used for classification to vary relative to each datapoint; data points that are easy to classify can be inferred on without looking at very many features, and more difficult datapoints can be classified using more features.
The underlying motivation is that, while classical approaches balance between accuracy and sparsity at the dataset level, our approach optimizes this balance at the individual datum level, thus resulting in equivalent accuracy at higher overall sparsity.
This kind of sparsity is interesting for several reasons: First, simpler explanations are always to be preferred as per Occam's Razor. Second, in the knowledge extraction process, such datum-wise sparsity is able to provide unique information about the underlying structure of the data space. Typically, if a dataset is organized onto two different subspaces, the datum-wise sparsity principle will allows the model to automatically choose to classify using only the features of one or another of the subspace.

DWSC considers feature selection and classification as a single sequential decision process. The classifier iteratively chooses which features to use for classifying each particular datum. In this sequential decision process, datum-wise sparsity is obtained by introducing a penalizing reward when the agent chooses to incorporate an additional feature into the decision process. The model is learned using an algorithm inspired by Reinforcement Learning \cite{SB}.

The contributions of the paper are threefold: (\textbf{i.}) We propose a new approach where classification is seen as a sequential process where one has to choose which features to use depending on the input being inferred upon. (\textbf{ii.}) This new approach results in a model that obtains good performance in terms of classification while maximizing datum-wise sparsity, i.e. the mean number of features used for classifying the whole dataset. It also naturally handles multi-class classification problems, solving them by using as few features as possible for all classes combined. (\textbf{iii.}) We perform a series of experiments on 14 different corpora and compare the model with those obtained by the LARS \cite{lars}, and a  $L_1$-regularized SVM, thus providing a qualitative study of the behaviour of our algorithm.

The paper is organized as follow: First, we define the notion of
\textbf{datum-wise sparse classifiers} and explain the interest of
such models in Section \ref{seq:dwsc}. We then describe our sequential approach to
classification and detail the learning algorithm and the complexity of
such an algorithm in Section \ref{part:seqseq}. We describe how this approach can be extended to
multi-class classification in Section \ref{sec:constrained}. We detail experiments on 14 datasets,
and also give a qualitative analysis of the behaviour of this model in Section \ref{sec:expe}. The related work is given in Section \ref{sec:relatedWork}.

\section{Datum-Wise Sparse Classifiers}
\label{seq:dwsc}
We consider the problem of supervised multi-class
classification\footnote{Note that this includes the binary supervised
  classification problem as a special case.} where one wants to learn
a classification function $f_\theta : \mathcal{X} \rightarrow \mathcal{Y}$ to associate one category $y \in \mathcal{Y}$ to a vector $\mathbf{x} \in \mathcal{X}$, where $\mathcal{X} = \mathbb{R}^n$, $n$ being the dimension of the input vectors.  $\theta$ is the set of parameters learned from a training set composed of input/output pairs $\mathcal{T}_{rain} = \{(\mathbf{x_i},y_i)\}_{i \in [1..N]}$.  These parameters are commonly found by minimizing the empirical risk defined by:

\begin{equation}
\theta^*=\argmin_\theta \frac{1}{N}\sum\limits_{i=1}^{N} \Delta(f_\theta(\mathbf{x_i}),y_i),
\label{eq:loss}
\end{equation}
where $\Delta$ is the loss associated to a prediction error.\\

This empirical risk minimization problem does not consider any \textit{prior assumption} or constraint concerning the form of the solution and can result in overfitting models. Moreover, when facing a very large number of features, obtained solutions usually need to perform computations on all the features for classifying any input, thus negatively impacting the model's classification speed. We propose a different risk minimization problem where we add a penalization term that encourages the obtained classifier to classify using \textbf{on average} as few features as possible. In comparison to classical $L_0$ or $L_1$ regularized approaches where the goal is to constraint the number of features used at the dataset level, our approach performs sparsity at the datum level, allowing the classifier to use different features when classifying different inputs. This results in a \textbf{datum-wise sparse classifier} that, when possible, only uses a few features for classifying easy inputs, and more features for classifying difficult or ambiguous ones.

We consider a different type of classifier function that, in addition to predicting a label $y$ given an input $\mathbf{x}$, also provides information about which features have been used for classification. Let us denote $\mathcal{Z} = \{0;1\}^n$. We define a \textbf{datum-wise classification function} $f$ of parameters $\theta$ as:

\begin{equation*}
 f_\theta : \begin{cases}
 						 \mathcal{X} \rightarrow \mathcal{Y} \times \mathcal{Z} \\
 						 f_\theta(\mathbf{x}) = (y,\mathbf{z})
 						\end{cases},
\end{equation*}
where $y$ is the predicted output and $\mathbf{z}$ is a $n$-dimensional vector $\mathbf{z}=(z^1,...,z^n)$, where $z^i=1$ implies that feature $i$ has been taken into consideration for computing label $y$ on datum $\mathbf{x}$. By convention, we denote the predicted label as $y_\theta(\mathbf{x})$ and the corresponding $\mathbf{z}$-vector as $z_\theta(\mathbf{x})$. Thus, if $z_\theta^i(\mathbf{x})=1$, feature $i$ has been used for classifying $\mathbf{x}$ into category $y_\theta(\mathbf{x})$.

This definition of data-wise classifiers has two main advantages: First, as we will see in the next section, because $f_\theta$ can explain its use of features with $z_\theta(\mathbf{x})$, we can add constraints on the features used for classification.  This  allows us to encourage datum-wise sparsity which we define below. Second, while this is not the main focus of our article, analysis of $z_\theta(\mathbf{x})$ gives a qualitative explanation of how the classification decision has been made, which we study in Section \ref{sec:expe}. Note that the way we define datum-wise classification is an extension to the usual definition of a classifier.
\subsection{Datum-Wise Sparsity}
Datum-wise sparsity is obtained by adding a penalization term to the empirical loss defined in equation \eqref{eq:loss} that limits the average number of features used for classifying:
\begin{equation}
\theta^*=\argmin_\theta \frac{1}{N}\sum\limits_{i=1}^{N} \Delta(y_\theta(\mathbf{x_i}),y_i) + \lambda \frac{1}{N} \sum\limits_{i=1}^{N} \Vert z_\theta(\mathbf{x_i}) \Vert_0.
\label{eq:datawiseloss}
\end{equation}
The term $\Vert z_\theta(\mathbf{x_i}) \Vert_0$ is the $L_0$ norm
\footnote{The $L_0$ 'norm' is not a proper norm, but we will refer to it as the $L_0$ norm in this paper, as is common in the sparsity community.} 
of $z_\theta(\mathbf{x_i})$,
i.e. the number of features selected for classifying $\mathbf{x_i}$,
that is, the number of elements in $z_\theta(\mathbf{x_i})$ equal to 1. In the general case, the minimization of this new risk results in a classifier that \textit{on average} selects only a few features for classifying, but may use a different set of features w.r.t to the input being classified. 
\textit{We consider this to be the crux of the DWSC model}: the classifier takes each datum into consideration differently during the inference process.

\begin{figure}[htb]
\vspace{-1cm}
\begin{center}
\begin{tabular}{cc}
\hspace{-1cm}
\includegraphics[width=0.6\linewidth]{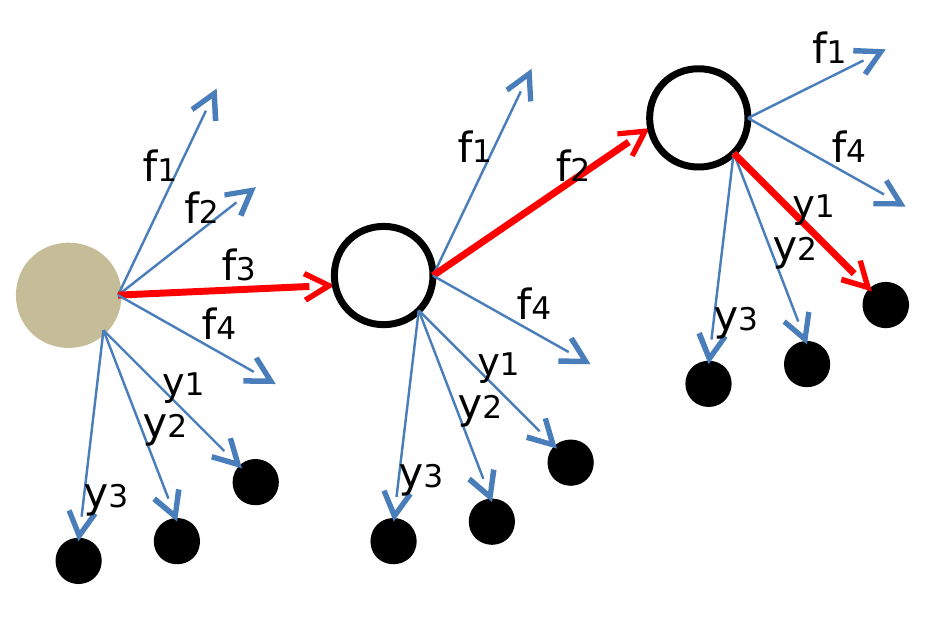} & \includegraphics[width=0.6\linewidth]{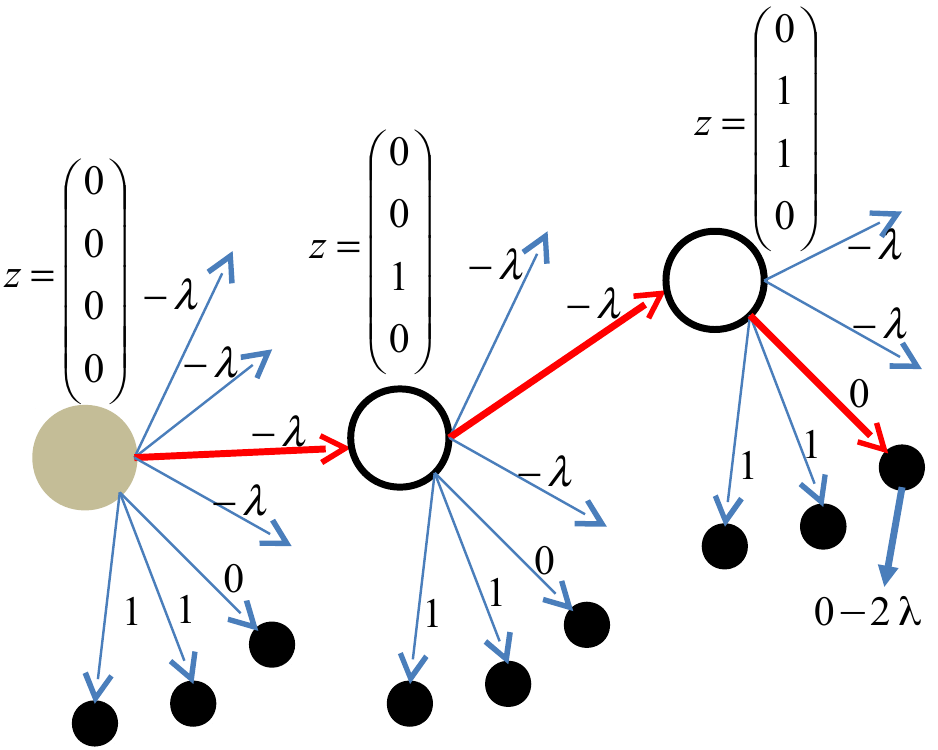} \\
(left)  & (right)
\end{tabular}
\end{center}
\caption{The sequential process for a problem with 4 features
  $(\mathbf{f_1},...,\mathbf{f_4})$ and 3 possible categories
  $(y_1,...,y_3)$. \textbf{Left:} The gray circle is the initial state
  for one particular input $\mathbf{x}$. Small circles correspond to
  terminal states where a classification decision has been made. In
  this example, the classification (bold arrows) has been made by
  sequentially choosing to acquire feature 3 then feature 2 and then
  to classify $\mathbf{x}$ in category $y_1$. The bold (red) arrows correspond to the trajectory made by the current policy. \textbf{Right: } The value of $z_\theta(\mathbf{x})$ for the different states are illustrated. The value on the arrows corresponds to the immediate reward received by the agent assuming that $\mathbf{x}$ belongs to category $y_1$. At the end of the process, the agent has received a total reward of $0-2\lambda$.}
\label{fig:fig1}
\end{figure}

Note that the optimization of the loss defined in equation \eqref{eq:datawiseloss} is a combinatorial problem that cannot be easily solved. In the next section of this paper, we propose an original way to deal with this problem, based on a Markov Decision Process.

\section{Datum-Wise Sparse Sequential Classification}
\label{part:seqseq}

We consider a Markov Decision Problem (MDP, \cite{Puterman1994})\footnote{The MDP is deterministic in our case.} to classify an input $\mathbf{x} \in \mathbb{R}^n$. At the beginning, we
have no information about $\mathbf{x}$, that is, we have no attribute/feature values. Then, at each step, we
can choose to acquire a particular feature of $\mathbf{x}$, or to
classify $\mathbf{x}$.  The act of classifying $\mathbf{x}$ in the
category $y$ ends an ``episode'' of the sequential process.
The classification process is a deterministic process defined by:
\begin{itemize}
\item A set of states $\mathcal{X} \times \mathcal{Z}$, where state
  $(\mathbf{x},\mathbf{z})$ corresponds to the state where the agent
  is currently classifying datum $\mathbf{x}$ and has selected
  features specified by $\mathbf{z}$. The number of currently selected features is thus $\Vert \mathbf{z} \Vert_0$.
\item A set of actions $\mathcal{A}$ where $\mathcal{A}(\mathbf{x},\mathbf{z})$ denotes the set of possible actions in state $(\mathbf{x},\mathbf{z})$. We consider two types of actions:
\begin{itemize}
\item $\mathcal{A}_f$ is the set of feature selection actions
  $\mathcal{A}_f= \{\mathbf{f_1}, \ldots, \mathbf{f_n}\}$ such that, for $a \in
  \mathcal{A}_f, a=\mathbf{f_j}$ corresponds to choosing feature $j$. Action
  $\mathbf{f_j}$ corresponds to a vector with only the $j^{\mbox{\scriptsize{}th}}$ element equal to
  1, i.e. $\mathbf{f_j}= (0, \ldots, 1, \ldots, 0)$.  Note that the set of possible
  feature selection actions on state $(\mathbf{x},\mathbf{z})$, denoted
  $\mathcal{A}_f(\mathbf{x},\mathbf{z})$, is equal to the subset of currently unselected
  features, i.e. $\mathcal{A}_f(\mathbf{x},\mathbf{z}) = \{\mathbf{f_j}, \text{ s.t. } z_j=0\}$.
\item  $\mathcal{A}_y$ is the set of classification actions
  $\mathcal{A}_y=\mathcal{Y}$, that correspond to assigning a label to the current
  datum. Classification actions stop the sequential decision process.
\end{itemize}
\item A transition function defined only for feature selection
  actions (since classification actions are terminal):
\[
\mathcal{T}:\begin{cases}
	\mathcal{X}\times\mathcal{Z}\times\mathcal{A}_f\rightarrow\mathcal{X}\times\mathcal{Z} \\
	\mathcal{T}((\mathbf{x},\mathbf{z}) , \mathbf{f_j}) = (\mathbf{x},\mathbf{z'})
	\end{cases}
\]
where $\mathbf{z'}$ is an updated version of $\mathbf{z}$ such that $\mathbf{z'}=\mathbf{z}+\mathbf{f_j}$.
\end{itemize}

\subsubsection{Policy}
\label{sec:policy}
We define a parameterized policy $\pi_\theta$, which, for each state $(\mathbf{x},\mathbf{z})$, returns the best action as defined by a scoring function $s_\theta(\mathbf{x},\mathbf{z},a)$:
\[
\pi_\theta : \mathcal{X} \times \mathcal{Z} \rightarrow \mathcal{A}\text{ and }\pi_\theta(\mathbf{x},\mathbf{z})=\argmax\limits_{a} s_\theta(\mathbf{x},\mathbf{z},a).
\]
The policy $\pi_\theta$ decides which action to take by applying the scoring function to every action possible from state $(\mathbf{x}, \mathbf{z})$ and greedily taking the highest scoring action.
The scoring function reflects the \textit{overall} quality of taking action $a$ in state $(\mathbf{x},\mathbf{z})$, which corresponds to the total reward obtained by taking action $a$ in $(\mathbf{x},\mathbf{z})$ and thereafter following policy $\pi_\theta$\footnote{This corresponds to the classical Q-function in Reinforcement Learning.}:
\[
	s_\theta(\mathbf{x},\mathbf{z},a) = r(\mathbf{x},\mathbf{z},a) + \sum\limits_{t=1}^{T} r_\theta^t \vert (\mathbf{x},\mathbf{z}),a.
\]

Here $(r_\theta^t \mid (\mathbf{x},\mathbf{z}),a)$ corresponds to the reward obtained at step $t$ while having started in state $(\mathbf{x},\mathbf{z})$ and followed the policy with parameterization $\theta$ for $t$ steps.  Taking the sum of these rewards gives us the total reward from state $(\mathbf{x},\mathbf{z})$ until the end of the episode.  Since the policy is deterministic, we may refer to a parameterized policy using simply $\theta$. Note that the optimal parameterization $\theta^*$
obtained after learning (see Sec.\@ \ref{sec:learning}) is the parameterization that maximizes the expected reward in all state-action pairs of the process.

In practice, the initial state of such a process for an input $\mathbf{x}$ corresponds to an empty $\mathbf{z}$ vector where no feature has been selected.  The policy $\theta$ sequentially picks, one by one, a set of features pertinent to the classification task, and then chooses to classify once enough features have been considered.

\subsubsection{Reward}
The reward function reflects the \textit{immediate} quality of taking
action $a$ in state $(\mathbf{x},\mathbf{z})$ relative
to the problem at hand.  We define a reward function over the training set $(\mathbf{x_i},y_i) \in T$: $\mathcal{R}:\mathcal{X}\times\mathcal{Z}\times\mathcal{A}\rightarrow \mathbb{R}$ which reflects how good of a decision taking action $\mathbf{f_j}$ on state $(\mathbf{x_i},\mathbf{z})$ for input $\mathbf{x_i}$ is relative to our classification task. This reward is defined as follows\footnote{Note that we can add $-\lambda \cdot \Vert \mathbf{z} \Vert_0$ to the reward at the end of the episode, and give a constant intermediate reward of $0$.  These two approaches are interchangeable.}:
\begin{itemize}
\item If $a$ corresponds to a feature selection action, then the reward is $-\lambda$.
\item If $a$ corresponds to a classification action i.e. $a=y$, we have:
\[
r(\mathbf{x_i},\mathbf{z},y) = 0 \text{ if }y = y_i \textbf{ and } = -1 \text{ if }y \neq y_i 
\]
\end{itemize}
In practice, we set $\lambda << 1$ to avoid situations where classifying incorrectly is a better decision than choosing multiple features.

\subsection{Reward Maximization and Loss Minimization}
As explained in section \ref{seq:dwsc}, our ultimate goal is to find the parameterization $\theta^*$ that minimizes the datum-wise empirical loss defined in equation \eqref{eq:datawiseloss}.  The training process for the MDP described above is the maximization of a reward function.  Let us therefore show that maximizing the reward function is equivalent to minimizing the datum-wise empirical loss.\\
\[
\begin{aligned}
	\theta^*&=  \argmin_\theta \frac{1}{N}\sum\limits_{i=1}^{N} \Delta(y_\theta(\mathbf{x_i}),y_i) + \lambda \frac{1}{N} \sum\limits_{i=1}^{N} ||z_\theta(\mathbf{x_i})||_0 \\
	&=  \argmin_\theta \frac{1}{N} \sum\limits_{i=1}^{N} \left( \Delta(y_\theta(\mathbf{x_i}),y_i) + \lambda  ||z_\theta(\mathbf{x_i})||_0 \right)\\
	&= \argmax_\theta \frac{1}{N} \sum\limits_{i=1}^{N} \left( -  \Delta(y_\theta(\mathbf{x_i}),y_i) - \lambda  ||z_\theta(\mathbf{x_i})||_0\right) \\
    &= \argmax_\theta \frac{1}{N} \sum\limits_{i=1}^{N} \begin{cases}
	0 - \lambda \cdot ||z_\theta(\mathbf{x_i})||_0 \text{ if }y = y_i\\
	-1 - \lambda \cdot ||z_\theta(\mathbf{x_i})||_0 \text{ if }y\neq y_i
\end{cases}\\
&=\argmax_\theta \frac{1}{N} \sum\limits_{i=1}^{N} \sum\limits_{t=1}^{T_\theta(\mathbf{x_i})+1} r(\mathbf{x_i}, z_\theta^{(t)}(\mathbf{x_i}), \pi_\theta(\mathbf{x_i}, z_\theta^{(t)}))
\end{aligned}
\]
where $\pi_\theta(\mathbf{x_i}, \mathbf{z}_\theta^{(t)})$ is the action taken
at time $t$ by the policy $\pi_\theta$ for the training example $\mathbf{x_i}$.

Such an equivalence between risk minimization and reward maximization
shows that the optimal classifier $\theta^*$ corresponds to the
optimal policy in the MDP defined previously. This equivalence allows
us to use classical MDP resolution algorithms in order to find the
best classifier. We detail the learning procedure in Section \ref{sec:learning}.

\subsection{Inference and Approximated Decision Processes}
\label{sec:project}

Due to the infinite number of possible inputs $\mathbf{x}$, the number of states is also infinite. Moreover, the reward function $r(\mathbf{x},\mathbf{z},a)$ is only known for the values of $\mathbf{x}$ that are in the training set and cannot be computed for any other input. For these two reasons, it is not possible to compute the score function for all state-action pairs in a tabular manner, and this function has to be approximated.

The scoring function that underlies the policy $s_\theta(\mathbf{x},\mathbf{z},a)$ is approximated with a linear model\footnote{ Although non-linear models such as neural networks may be used, we have chosen to restrict ourselves to a linear model to be able to properly compare performance with that of other state-of-the-art linear sparse models.}:
\[
s(\mathbf{x},\mathbf{z},a) = \langle \Phi(\mathbf{x},\mathbf{z},a) ; \theta \rangle
\] and the policy defined by such a function consists in taking in state $(\mathbf{x},\mathbf{z})$ the action $a'$ that maximizes the scoring function i.e $a' = \argmax_{a\in\mathcal{A}} \langle \Phi(\mathbf{x},\mathbf{z},a) ; \theta \rangle$.

Due to their infiniteness, the state-action pairs are represented in a
feature space. We note $\Phi(\mathbf{x},\mathbf{z},a)$ the featurized
representation of the $(\mathbf{x},\mathbf{z}),a$ state-action pair.
Many definitions may be used for this feature representation, but we propose
a simple projection:
we restrict the representation of $\mathbf{x}$ to only the selected
features. Let $\mu(\mathbf{x}, \mathbf{z})$ be the restriction of
$\mathbf{x}$ according to $\mathbf{z}$:
\[
	\mu(\mathbf{x},\mathbf{z})^i = \begin{cases}
		x^i\text{ if }z^i=1 \\
		0 \text{ elsewhere}
	\end{cases}.
\]


To be able to differentiate between an attribute of $\mathbf{x}$ that
is not yet known, and an attribute that is simply equal to 0, we must keep the information present in $\mathbf{z}$.  Let  $\phi(\mathbf{x},\mathbf{z}) = (\mathbf{z} , \mu(\mathbf{x},\mathbf{z}))$ be the intermediate representation that corresponds to the \textit{concatenation} of $\mathbf{x}$ with $\mathbf{z}$.  Now we simply need to keep the information present in $a$ in a manner that allows each action to be easily distinguished by a linear classifier.  To do this we use the block-vector trick \cite{Har-Peled2002} which consists in projecting $\phi(\mathbf{x}, \mathbf{z})$ into a higher dimensional space such that the position of $\phi(\mathbf{x}, \mathbf{z})$ inside the global vector $\Phi(\mathbf{x}, \mathbf{z}, ,a)$ is dependent on action $a$:
\[
\\ \Phi(\mathbf{x},\mathbf{z},a) = \left( 0, \ldots ,0, \Phi(\mathbf{x},\mathbf{z}), 0, \ldots, 0 \right).
\]
In $\Phi(\mathbf{x},\mathbf{z},a)$, the block $\phi(\mathbf{x},\mathbf{z})$ is at position $i_a \cdot |\phi(\mathbf{x},\mathbf{z})|$ where $i_a$ is the index of action $a$ in the set of all the possible actions. Thus, $\phi(\mathbf{x}, \mathbf{z})$ is offset by an amount dependent on the action $a$.


\subsection{Learning}
\label{sec:learning}
The goal of the learning phase is to find an optimal policy
parameterization $\theta^*$ which maximizes the expected reward, thus
minimizing the datum-wise regularized loss defined in
\eqref{eq:datawiseloss}. As explained in Section \ref{sec:project}, we
cannot exhaustively explore the state space during training, and
therefore we use a Monte-Carlo approach to sample example states from the learning space.
We use the Approximate Policy Iteration (API) algorithm with
rollouts \cite{Lagoudakis2003}. Sampling state-action pairs according
to a previous policy $\pi_{\theta^{(t-1)}}$,
API consists in iteratively learning a better policy
$\pi_{\theta^{(t)}}$ by way of the Bellman equation. The API With Rollouts algorithm is composed of three main steps that are iteratively repeated:
\begin{enumerate}
 \item The algorithm begins by sampling a set of random states: the $\mathbf{x}$ vector is sampled from a uniform distribution in the training set, and $\mathbf{z}$ is also sampled using a uniform binomial distribution.
 \item For each state in the sampled state, the policy $\pi_{\theta^{(t-1)}}$ is used to compute the expected reward of choosing each possible action from that state. We now have a feature vector $\Phi(\mathbf{x},\mathbf{z},a)$ for each state-action pair in the sampled set, and the corresponding expected reward denoted $R_{\theta^{(t-1)}}(\mathbf{x},\mathbf{z},a)$.
 \item The parameters $\theta^{(t)}$ of the new policy are then computed using classical linear regression on the set of states --- $\Phi(\mathbf{x},\mathbf{z},a)$ --- and corresponding expected rewards --- $R_{\theta^{(t-1)}}(\mathbf{x},\mathbf{z},a)$ --- obtained previously.  The generalizing capacity of the classifier gives an estimated score to state-action pairs even if we have never visited them.
\end{enumerate}
After a certain number of iterations, the parameterized policy converges to a final policy $\pi$ which is used for inference.

\section{Preventing Overfitting in the Sequential Model}
\label{sec:constrained}
In section \ref{sec:policy}, we explain the process by which, at each
step, we either choose a new feature or classify the current datum. This
process is at the core of DWSC but can suffer from overfitting if the
number of features is larger than the number of training examples. In
such a case, DWSC would tend to learn to select the more specific
features for each training example. In classical $L_1$ regularization
models that are not datum-wise, the classifier must use the same set
of features for classifying any data and thus overly specific features
are not chosen because they usually appear in only a few training
examples.

We propose a very simple variant of the general model that allows us
to avoid overfitting. We still allow DWSC to choose how many features
to use before classifying an input $x$, but we constrain it to choose
the features \textbf{in the same order} for all the inputs.  For that, we constrain the score of the feature
selection actions to depend only on the vector $\mathbf{z}$ of the
state $(\mathbf{x},\mathbf{z})$.  An example of the effect of such a constraint is presented in Fig.\@
\ref{fig:fig3}.  This constraint is handled in the
following manner:
\begin{equation}
\forall (\mathbf{x},\mathbf{z},a) \begin{cases}
	s_\theta(\mathbf{x},\mathbf{z},a) = s_\theta(\mathbf{z},a) \text{ if } a \in \mathcal{A}_f \\
	s_\theta(\mathbf{x},\mathbf{z},a) = s_\theta(\mathbf{x},\mathbf{z},a) \text{ if } a \in \mathcal{A}_y
	\end{cases},
\end{equation}
where $s_\theta(\mathbf{x},\mathbf{z},a) = s_\theta(\mathbf{z},a)$ implies that the score is computed using only the values of $\mathbf{z}$ and $a$ --- $\mathbf{x}$ is ignored. This corresponds to having two different types of state-action feature vectors $\Phi$ depending on the type of action:
\begin{equation}
\begin{cases} \text{if } a\in \mathcal{A}_f, \Phi(\mathbf{x},\mathbf{z},a)=\left( 0, \ldots, 0,  \mathbf{z},  0, \ldots, 0\right) \\
\text{if } a\in\mathcal{A}_y,  \Phi(\mathbf{x},\mathbf{z},a)= \left( 0, \ldots, 0,  \mathbf{z}, \Phi(\mathbf{x},\mathbf{z}), 0, \ldots, 0 \right)
\end{cases}.
\end{equation}

\begin{figure}[h]
\vspace{-0.5cm}
\begin{center}
\small{
\begin{tabular}{||c|p{1cm}p{1cm}p{1cm}p{1cm}|||c|p{1cm}p{1cm}p{1cm}p{1cm}||} \hline \hline
Example & \multicolumn{4}{l|||}{Features Selected} & Example & \multicolumn{4}{l||}{Features Selected} \\ \hline
$\mathbf{x_1}:$ & 2 & 3 & & & $\mathbf{x_1}$: & 2 & 3 & &  \\
$\mathbf{x_2}:$ & 1 & 4 & 2 & 3 & $\mathbf{x_2}$: & 2 & 3 & 1 & 4 \\
$\mathbf{x_3}:$ & 3 & & & & $\mathbf{x_3}$: & 2 & 3 & &  \\
$\mathbf{x_4}:$ & 2 & 3 & 1 & & $\mathbf{x_4}$: & 2 & 3 & 1 &  \\ \hline
\multicolumn{5}{||c|||}{Unconstrained Model} & \multicolumn{5}{c||}{Constrained Model} \\ \hline
\end{tabular}
}
\end{center}
\caption{Difference between the base \textit{Unconstrained Model} (DWSM-Un) and the \textit{Constrained Model} (DWSM-Con) described in section \ref{sec:constrained}. The figure shows, for 4 different inputs $\mathbf{x_1},...,\mathbf{x_4}$ the features selected by the classifiers before classification. One can see that the \textit{Constrained Model} chooses the features in the same order for all the inputs.}
\label{fig:fig3}
\end{figure}

Although this constraint forces DWSC to choose the features in the same order, it will still automatically learn the best order in which to choose the features, and when to stop adding features and classify.  However, it will avoid choosing very different features sets for classifying different inputs (the first features chosen will be common to all the inputs being classified) and thus avoid the overfitting problem.

\section{Complexity Analysis}
\vspace{-0.35cm}
\paragraph{Learning Complexity: } As explained in section \ref{sec:learning}, the learning method is based on Reinforcement Learning with Rollouts. Such an approach is expensive in term of computations because it needs --- at each iteration of the algorithm --- to simulate trajectories in the decision process, and then to learn the scoring function $s_\theta$ based on these trajectories. Without giving the details of the computation, the complexity of each iteration is $O(N_s \cdot (n^2+c))$, where $N_s$ is the number of states used for rollouts (which in practice is proportional to the number of training examples), $n$ is the number of features and $c$ is the number of possible categories. This implies a learning method which is quadratic w.r.t. the number of features; the proposed approach is not able to deal with problems with thousands of possible features. Breaking this complexity is an active research perspective with some leads.
\paragraph{Inference Complexity: } Inference on an input $\mathbf{x}$ consists in sequentially choosing features, and then classifying $\mathbf{x}$. At step $t$, one has to perform  $(n-t)+c$ linear computations in order to choose the best action, where $(n-t)+c$ is the number of possible actions when $t$ features have already been acquired. The inference complexity is thus $O(N_f \cdot (n+c))$, where $N_f$ is the mean number of features chosen by the system before classifying. In fact, due to the shape of the $\Phi$ function presented in Section \ref{sec:project} and the linear nature of $s_\theta$, the score of the actions can be efficiently incrementally computed at each step of the process by just adding the contribution of the newly added feature. The complexity is thus reduced to $O(n+c)$. Moreover, the constrained model which results in ordering the features, has a lower complexity of $O(c)$ because in that case, the model does not have to choose between the different remaining features, and has only the choice to classify or get the next feature w.r.t. to the learned order.\\
\linebreak
If the learning complexity of our model is higher than baseline global linear methods, the inference speed is very close for the unconstrained model, and equivalent for the constrained one.  In practice, most of the baseline methods choose a subset of variables in a couple seconds to a couple minutes, whereas our method takes from a dozen minutes to an hour, depending on the number of features and categories.  In practice inference is indeed of the same speed, which is in our opinion the important factor.

\section{Experiments}
\vspace{-0.35cm}
\label{sec:expe}

\begin{figure}[t]
\vspace{-1cm}
\begin{center}
\begin{tabular}{cc}
\hspace{-1.5cm} \includegraphics[width=0.7\linewidth]{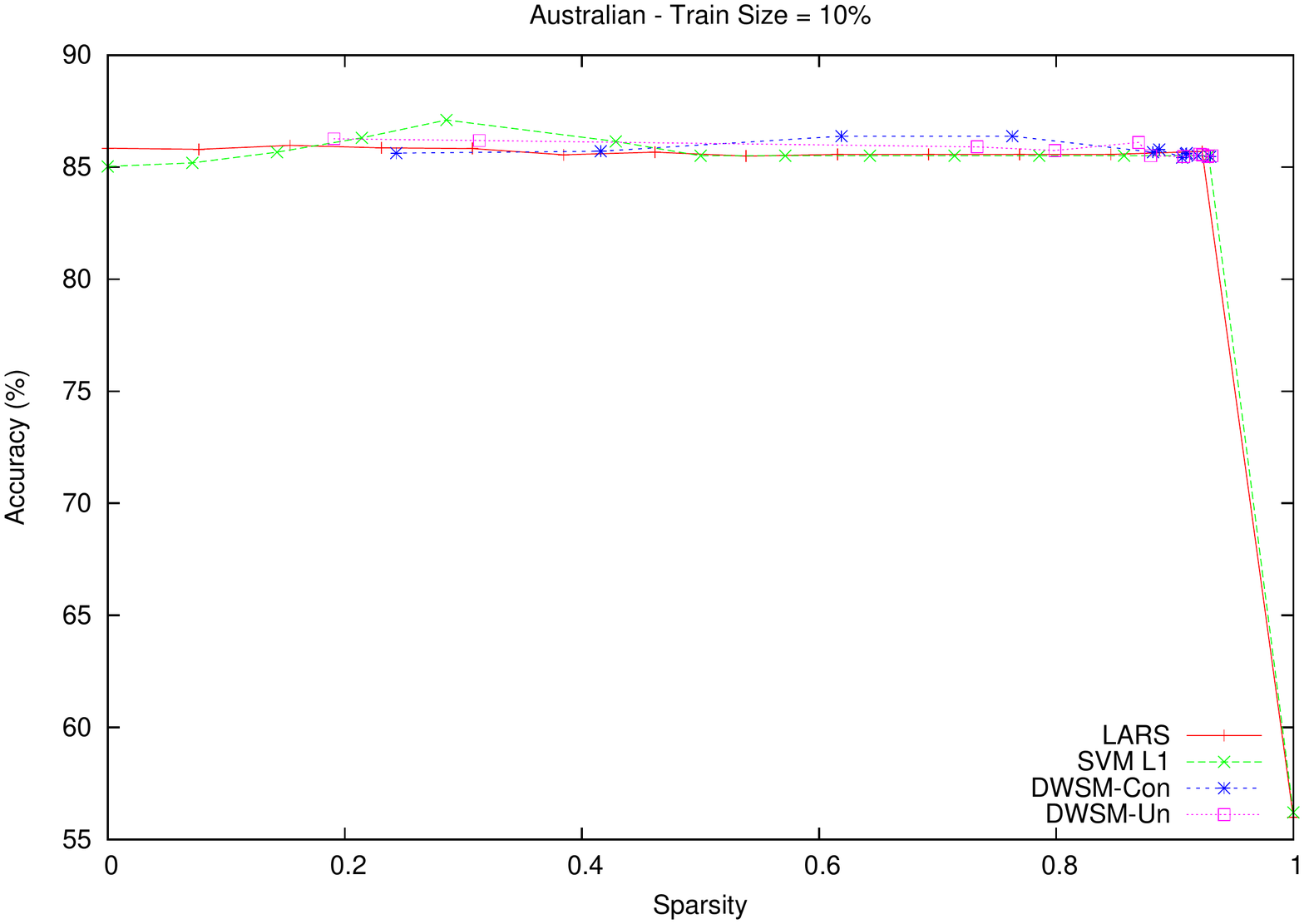} & \hspace{-1.5cm}  \includegraphics[width=0.7\linewidth]{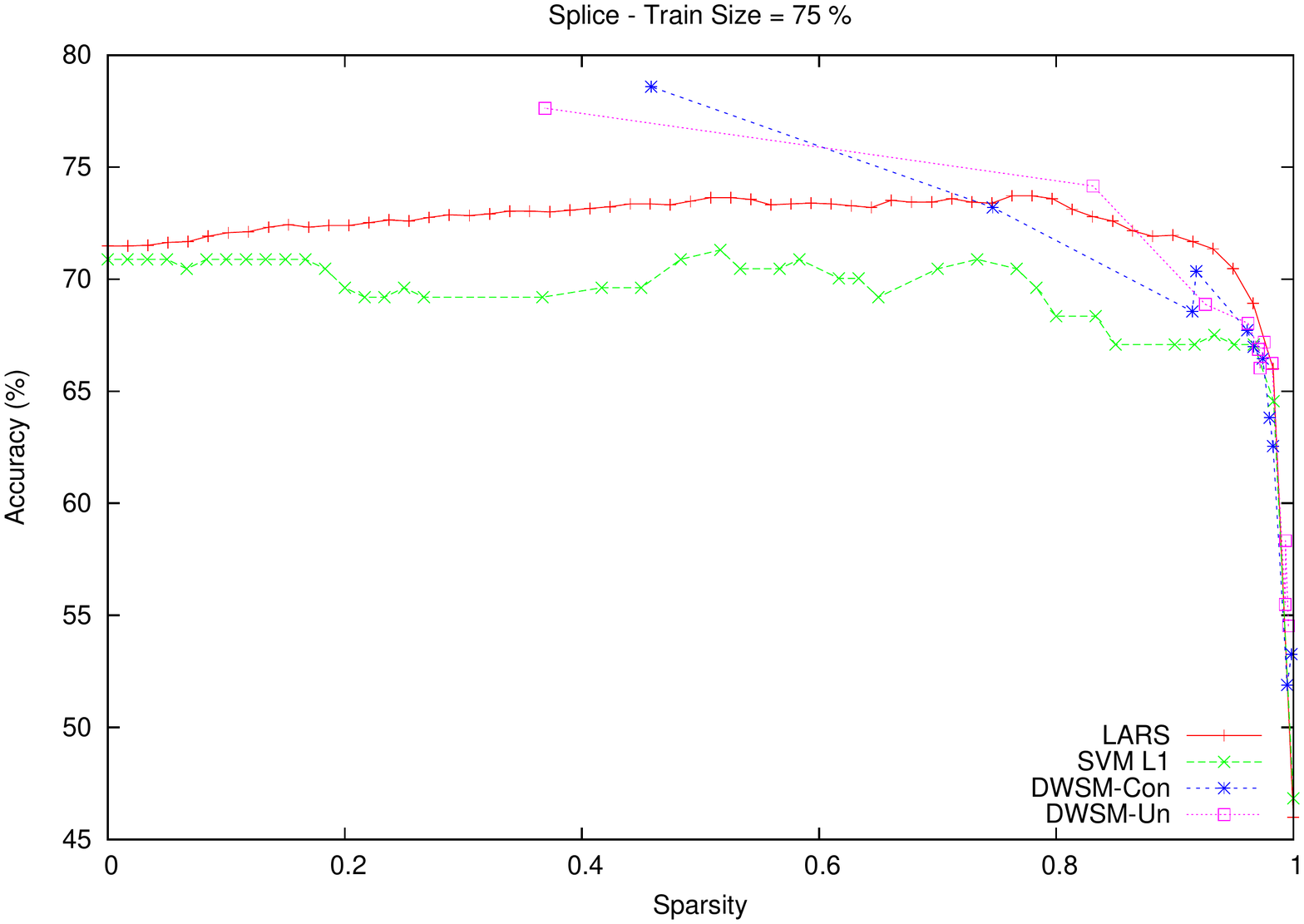}
\end{tabular}
\end{center}
\caption{Accuracy w.r.t.\@ to sparsity. In both plots, the left side
  on the x-axis corresponds to a low sparsity, while the right side corresponds to a high sparsity. The performances of the models are usually decreasing when the sparsity increases, except in case of overfitting.}
\label{fig:curve1}
\end{figure}

\begin{table}[t]
\begin{center}
\tiny{
\begin{tabular}{|l|c|c|c||c|} \hline
Name & Number of Features & Number of examples & Number of Classes & Task \\ \hline \hline
Australian  & 14 & 690 & 2 & Binary \\
Breast Cancer  & 10 & 683 & 2 & Binary \\
Diabetes  &  8 & 768 & 2 & Binary \\
German Numer  & 24 & 1,000 & 2 & Binary \\
Heart & 13 & 270 & 2 & Binary \\
Ionosphere & 34 & 351 & 2 & Binary \\
Liver Disorders & 6 & 345 & 2 & Binary \\
Sonar  & 60 & 208 & 2 & Binary \\
Splice  & 60 & 1,000 & 2 & Binary \\
Svm Guide 3 & 21 & 1,284 & 2 & Binary \\ \hline
Segment  & 19 & 2,310 & 7 & Multiclass \\
Vehicle & 18 & 846 & 4 & Multiclass \\
Vowel & 10 & 1,000 & 11 & Multiclass \\
Wine & 13 & 178 & 3 & Multiclass \\ \hline
\end{tabular}
}
\end{center}
\caption{Datasets used for the experiments.}
\label{tab:uci}
\end{table}


Experiments were run on 14 different datasets obtained from the LibSVM Website\footnote{\url{http://www.csie.ntu.edu.tw/~cjlin/libsvmtools/datasets/}}. Ten of these datasets correspond to a binary classification task, four to a multi-class problem. The datasets are described in Table \ref{tab:uci}.  For each dataset, we randomly sampled different training sets by taking from $5\%$ to $75\%$ of the examples as training examples, with the remaining examples being kept for testing. We performed experiments with three different models: \textbf{L1-SVM} was used as a baseline linear model with $L_1$ regularization\footnote{Using LIBLINEAR \cite{Fan2008}.}. \textbf{LARS} was used to obtain the optimal solution of the LASSO problem for all values of the regularization coefficient $\lambda$ at once\footnote{We use the implementation from the authors of the LARS, available in R.}. \textbf{Datum-Wise Sequential Model (DWSM)} was tested with the two versions presented above: (i) \textbf{DWSM-Un} is the original unconstrained model and (ii) \textbf{DWSM-Con} is the constrained model for preventing overfitting.

For the evaluation, we used a classical accuracy measure which corresponds to $1-\textit{error rate}$ on the test set of each dataset. We perform 3 training/testing set splits of a given dataset to obtain averaged figures. The sparsity has been measured as the proportion of features \textit{not} used for $L_1$-SVM and LARS in binary classification, and the \textit{mean} proportion of features not used to classify \textbf{testing examples} in DWSM. For multi-class problems where one LARS/SVM model is learned for each category, the sparsity is the proportion of features that have not been used in \textit{any} of the models.

For the sequential experiments, the number of rollout states (step 1 of the learning algorithm) has been set to 2,000 and the number of policy iterations has been fixed to 10. Note that experiments with more rollout states and/or more iterations give similar results. Experiments were made using an \textit{alpha mixture policy} with $\alpha=0.9$ to ensure the stability of the learning process. We tested the different models with different values of $\lambda$ which controls the sparsity. Note that even with a $\lambda=0$ value, contrary to the baseline models, the DWSM model does not use all of the features for classification.

\begin{table}[t]
\tiny{
\begin{center}
\begin{tabular}{|c|c||ccc|ccc|ccc|} \hline
Corpus & Train Size & \multicolumn{3}{||c|}{Sparsity = 0.8}& \multicolumn{3}{|c|}{Sparsity = 0.6}& \multicolumn{3}{|c|}{Sparsity = 0.4} \\ \hline
 & & DWSM-Un & DWSM-Con & SVM L1 & DWSM-Un & DWSM-Con & SVM L1  & DWSM-Un & DWSM-Con & SVM L1  \\ \hline
\multirow{4}{1.5cm}{australian}
& 0.05 & 85.31 & 85.13  & \textbf{85.52} &  84.63 & 84.83 & \textbf{85.42}  & 84.01 & 84.32 & \textbf{85.15} \\
& 0.1 & 85.75  & \textbf{86.16} &  85.51 &  86.00 & \textbf{86.32} &  85.51  & 86.13 & 85.70 & \textbf{86.34} \\
& 0.25 & 85.39  & \textbf{86.16}  & 85.33 &  86.76 & \textbf{86.99}  & 85.33  & \textbf{86.56} & 86.49 & 86.10 \\
& 0.5 & 83.58  & \textbf{84.13}  & 83.47 &  \textbf{84.33}  & 84.19&  83.47  & \textbf{84.85} & 84.28 & 83.70 \\ \hline
\multirow{4}{1.5cm}{breast-cancer}
& 0.05 & 87.73  & \textbf{88.50} & 88.17 & \textbf{96.58}  & 96.49 &  94.93 & 96.57 & \textbf{96.68} & 96.62 \\
& 0.1  & \textbf{89.25}  & 89.16 & 88.19 & 94.70 & \textbf{95.13}  & 91.91 & 95.42 & \textbf{95.63} & 92.88 \\
& 0.25 & 91.16 & 88.48  & \textbf{92.64}  & \textbf{97.06}  & 96.76 & 94.38 & 97.09 & \textbf{97.11} & 96.71 \\
& 0.5 & 82.65 & 82.60  & \textbf{90.26}  & \textbf{96.92}  & 95.98 & 93.41 & 96.50 & \textbf{97.01} & 94.84 \\ \hline
\multirow{4}{1.5cm}{diabetes}
& 0.05 & 68.61  & \textbf{69.10}  & 65.85 & 71.30 & \textbf{71.36}  & 67.46 & \textbf{72.15} & 0.00 & 71.07 \\
& 0.1  & \textbf{70.92}  & 69.14 & 65.71 & \textbf{72.52}  & 71.62 & 64.97 & \textbf{73.26} & 72.81 & 70.23 \\
& 0.25 & 68.39  & \textbf{68.58}  & 64.88 & 71.83 & \textbf{72.42}  & 65.88 & 74.80 & 74.96 & \textbf{75.09} \\
& 0.5  & \textbf{70.65}  & 69.69 & 62.27 & \textbf{72.67}  & 70.90 & 67.30 & \textbf{73.82} & 73.62 & 72.14 \\ \hline
\multirow{4}{1.5cm}{german.numer}
& 0.05  & \textbf{70.58}  & 70.47 & 67.36 & \textbf{70.74}  & 70.39 & 69.95 & 69.99 & \textbf{70.28} & 69.73 \\
& 0.1  & \textbf{69.82}  & 69.62 & 69.10 & 70.81 & 70.39 & \textbf{71.52}  & 71.79 & 0.00 & \textbf{72.85} \\
& 0.25  & \textbf{72.25}  & 72.00 & 65.98 & 72.67 & \textbf{73.26}  & 72.89 & 73.10 & 0.00 & \textbf{74.11} \\
& 0.5 & 70.03  & \textbf{70.62}  & 69.72 & 71.50 & \textbf{72.37}  & 71.97 & 72.96 & \textbf{74.05} & 72.68 \\ \hline
\multirow{4}{1.5cm}{heart}
& 0.05  & \textbf{48.33}  & 48.17 & 45.42 & 51.17 & 50.67 & \textbf{65.73}  & 0.00 & 0.00 & \textbf{68.24} \\
& 0.1  & \textbf{75.50}  & 74.27 & 73.42 & \textbf{76.61}  & 75.78 & 73.76 & \textbf{77.60} & 77.49 & 74.94 \\
& 0.25 & 76.17  & \textbf{78.50}  & 76.26 & 81.31 & 81.70 & \textbf{83.33}  & 82.24 & 83.00 & \textbf{83.64} \\
& 0.5  & \textbf{70.34}  & 68.87 & 69.20 & 77.15 & \textbf{80.34}  & 78.83 & 80.48 & 80.40 & \textbf{80.58} \\ \hline
\multirow{4}{1.5cm}{ionosphere}
& 0.05 & 69.52 & 71.36  & \textbf{73.55}  & \textbf{73.44}  & 73.02 & 72.23 & 74.77 & \textbf{75.16} & 72.59 \\
& 0.1 & 71.58 & 71.09  & \textbf{71.84}  & 75.12 & 74.63 & \textbf{75.89}  & \textbf{74.97} & 74.93 & 74.49 \\
& 0.25 & 79.65  & \textbf{80.29}  & 75.94 & 85.18 & \textbf{85.44}  & 81.58 & 85.58 & \textbf{85.69} & 82.78 \\
& 0.5 & 77.31  & \textbf{78.40}  & 71.15 & \textbf{82.94}  & 82.68 & 78.18 & \textbf{84.96} & 84.16 & 79.64 \\ \hline
\multirow{4}{1.5cm}{liver-disorders}
& 0.05  & \textbf{60.40}  & 59.37 & 57.01 & 60.07 & \textbf{61.25}  & 57.01 & 60.29 & \textbf{64.27} & 57.74 \\
& 0.1  & \textbf{56.70}  & 56.24 & 55.41 & 55.85 & 55.98 & \textbf{56.43}  & \textbf{56.69} & 55.00 & 55.86 \\
& 0.25  & \textbf{56.69}  & 56.14 & 54.18 & \textbf{58.07}  & 57.02 & 55.10 & \textbf{58.69} & 57.97 & 56.93 \\
& 0.5 & 58.93 & 59.55  & \textbf{60.84}  & 60.10 & 58.81 & \textbf{60.96}  & 59.33 & 60.84 & \textbf{61.33} \\ \hline
\multirow{4}{1.5cm}{sonar}
& 0.05 & 57.59 & 59.95  & \textbf{64.14}  & \textbf{68.50}  & 66.49 & 65.15 & 69.45 & \textbf{70.48} & 61.24 \\
& 0.1 & 61.69  & \textbf{64.40}  & 64.12 & 68.68 & \textbf{73.93}  & 64.12 & 74.25 & \textbf{75.20} & 63.53 \\
& 0.25 & 67.32 & 64.74  & \textbf{67.52}  & 73.52 & 70.63 & \textbf{74.52}  & \textbf{75.22} & 73.36 & 72.82 \\
& 0.5  & \textbf{68.19}  & 64.71 & 65.77 & \textbf{72.18}  & 69.76 & 69.37 & \textbf{73.73} & 71.60 & 65.77 \\ \hline
\multirow{4}{1.5cm}{splice}
& 0.05 & 67.23  & \textbf{68.41}  & 67.82 & \textbf{70.14}  & 68.66 & 65.93 & \textbf{70.51} & 69.89 & 64.47 \\
& 0.1  & \textbf{66.90}  & 66.87 & 61.46 & \textbf{70.35}  & 67.99 & 62.63 & \textbf{71.05} & 70.07 & 61.62 \\
& 0.25 & 73.87  & \textbf{73.89}  & 70.49 & 74.81 & \textbf{75.30}  & 72.28 & 75.60 & \textbf{76.64} & 71.74 \\
& 0.5 & 72.86  & \textbf{76.79}  & 72.78 & 74.98 & \textbf{77.88}  & 70.36 & \textbf{77.09} & 0.00 & 69.35 \\ \hline
\multirow{4}{1.5cm}{svmguide3}
& 0.05 & 77.15 & 77.13  & \textbf{77.17}  & 77.32 & 77.25 & \textbf{78.25}  & 77.48 & 77.37 & \textbf{78.21} \\
& 0.1  & \textbf{77.31}  & 77.28 & 76.59 & 77.94 & 78.11 & \textbf{78.95}  & 78.58 & \textbf{78.94} & 78.37 \\
& 0.25  & \textbf{76.67}  & 76.56 & 75.96 & \textbf{77.44}  & 77.14 & 77.40 & \textbf{78.21} & 77.72 & 77.91 \\
& 0.5 & 77.71  & \textbf{77.78}  & 76.87 & 78.55 & \textbf{78.63}  & 78.15 & 79.38 & \textbf{79.47} & 78.37 \\ \hline
\end{tabular}
\end{center}
\caption{This table contains the accuracy of each model on the binary classification problems depending on three levels of sparsity ($80\%$, $60\%$, and $40\%$) using different training sizes. The accuracy has been linearly interpolated from curves like the ones given in Figure \ref{fig:curve1}.}
\label{table:resbinary}
}
\vspace{-0.55cm}
\end{table}

\vspace{-0.35cm}
\subsection{Results}
\vspace{-0.25cm}
For each corpus and each training size, we have computed sparsity/accuracy curves showing the performance of the different models w.r.t. to the sparsity of the solution. Only two representative curves are given in Figure \ref{fig:curve1}. To summarize the performances over all the datasets, we give the accuracy of the different models for three levels of sparsity in tables \ref{table:resbinary} and \ref{table:resmulticlass}. Due to a lack of space, these tables do not present the LARS' performance, which  are equivalent to the performances of the $L_1$-SVM. Note that in order to obtain the accuracy for a given level of sparsity, we have computed a linear interpolation on the different curves obtained for each corpus and each training size.  This linear interpolation allows us to compare the baseline sparsity methods --- that choose a fixed number of features --- with the average number of features chosen by DWSC.  This compares the average amount of information considered by each classifier.  We believe this approach still provides a good appreciation of the algorithm's capacities.\\
\begin{table}[t]
\tiny{
\begin{center}
\begin{tabular}{|c|c||ccc|ccc|ccc|} \hline
Corpus & Train Size & \multicolumn{3}{||c|}{Sparsity = 0.8}& \multicolumn{3}{|c|}{Sparsity = 0.6}& \multicolumn{3}{|c|}{Sparsity = 0.4} \\ \hline
 & & DWSM-Un & DWSM-Con & L1-SVM & DWSM-Un & DWSM-Con & L1-SVM & DWSM-Un & DWSM-Con & L1-SVM  \\ \hline
\multirow{4}{1.5cm}{segment}
& 0.1  & \textbf{42.06}  & 41.23 & 35.31 & \textbf{53.87}  & 53.02 & 45.49 & 54.83 & 56.57 & \textbf{56.98} \\
& 0.2  & \textbf{40.76}  & 40.17 & 40.48 & 55.70 & \textbf{56.34}  & 45.97 & 57.42 & \textbf{59.10} & 53.24 \\
& 0.5  & \textbf{43.29}  & 0.00 & 37.17 & \textbf{54.09}  & 0.00 & 45.15 & \textbf{56.43} & 0.00 & 50.52 \\
& 0.75  & \textbf{43.78}  & 41.13 & 38.22 & \textbf{55.10}  & 53.60 & 44.80 & 56.54 & \textbf{56.99} & 47.00 \\ \hline
\multirow{4}{1.5cm}{vehicle}
& 0.1 & 34.23 & 37.52  & \textbf{43.36}  & 43.50 & 45.34 & \textbf{50.25}  & 47.21 & 0.00 & \textbf{56.54} \\
& 0.2 & 38.32 & 39.27  & \textbf{53.04}  & 45.84 & 45.68 & \textbf{53.36}  & 48.68 & 47.91 & \textbf{52.83} \\
& 0.5 & 39.74 & 39.51  & \textbf{42.95}  & 46.64 & 47.57 & \textbf{50.30}  & 0.00 & 48.40 & \textbf{51.99} \\
& 0.75 & 40.32 & 40.37  & \textbf{41.04}  & 49.96 & 49.31 & \textbf{53.68}  & 51.86 & 51.53 & \textbf{53.77} \\ \hline
\multirow{4}{1.5cm}{vowel}
& 0.1 & 18.03  & \textbf{19.27}  & 9.83 & \textbf{24.17}  & 22.82 & 16.24 & 25.28 & \textbf{25.80} & 18.38 \\
& 0.2 & 0.00  & \textbf{15.27}  & 14.71 &  & \textbf{20.17}  & 15.93 & 0.00 & \textbf{22.59} & 15.93 \\
& 0.5  & \textbf{18.98}  & 17.81 & 9.57 & 24.56 & \textbf{25.33}  & 17.73 & \textbf{28.45} & 27.31 & 23.76 \\
& 0.75  & \textbf{19.85}  & 19.49 & 14.41 & 28.01 & \textbf{31.45}  & 24.58 & 32.09 & \textbf{32.74} & 26.69 \\ \hline
\multirow{4}{1.5cm}{wine}
& 0.1 & 70.22 & 70.66  & \textbf{73.58}  & 76.42 & 77.87 & \textbf{89.38}  & 78.66 & 76.67 & \textbf{91.36} \\
& 0.2 & 71.52 & 72.68  & \textbf{80.34}  & 78.27 & 79.11 & \textbf{92.12}  & 78.76 & 77.72 & \textbf{94.16} \\
& 0.5 & 72.99 & \textbf{74.41}  & 74.40  & 79.43 & 80.60 & \textbf{86.90}  & 82.15 & 79.50 & \textbf{91.38} \\
& 0.75  & \textbf{76.21}  & 75.04 & 72.00 & 80.18 & 81.84 & \textbf{94.00}  & 83.23 & 80.93 & \textbf{96.00} \\ \hline
\end{tabular}
\end{center}
\caption{This table contains the accuracy of each model on the multi-class classification problems depending on three levels of sparsity ($80\%$, $60\%$, and $40\%$) using different training sizes.}
\label{table:resmulticlass}
}
\end{table}

Table \ref{table:resbinary} shows that, for a sparsity level of 80\%, the DWSM-Un and the DWSM-Con models outperform the baseline $L_1$-SVM classifier. This is particularly true for 7 of the 10 datasets while the results are more ambiguous on the three others datasets: \textit{breast, ionosphere and sonar}. For a sparsity of 40\%, similar results are obtained.  Depending on the corpus and the training size, different configurations are observed. Some datasets can be easily classified using only a few features, such as \textit{australian} for example. In that case, our approach gives similar results in comparison to $L_1$ methods (see Figure \ref{fig:curve1}--left). For some other datasets, our method clearly outperforms baseline methods (Figure \ref{fig:curve1}--right). On the \textit{splice} dataset, our model is better than the best (non-sparse) SVM using only less than 20\% of the features on average. This is due to the fact that our sequential process, which solves a different classification problem, is more appropriate for some particular datasets, particularly when the distribution of the data is split up amongst distinct subspaces.
In this case, our model is able to choose more appropriate features for each input.\\

When using small training sets with some datasets  --- \textit{sonar} or \textit{ionosphere} --- where
overfitting is observed (accuracy decreases with more features used), the DWSM-Con seems to be a better choice than the unconstrained version and thus is a version of the algorithm that is well-suited when the number of learning examples is small.

Concerning the multi-class problems, similar effects can be observed (see Table \ref{table:resmulticlass}). The model seems particularly interesting when the number of categories is high, as in \textit{segment} and \textit{vowel}. This is due to the fact that the average sparsity is optimized by the sequential model for the multi-class problem while $L_1$-SVM and LARS, which need to learn one model for each category, perform separate sparsity optimizations for each class.
\begin{figure}
\begin{center}
\vspace{-1cm}
\begin{tabular}{cc}
\hspace{-1cm} \includegraphics[width=0.6\linewidth]{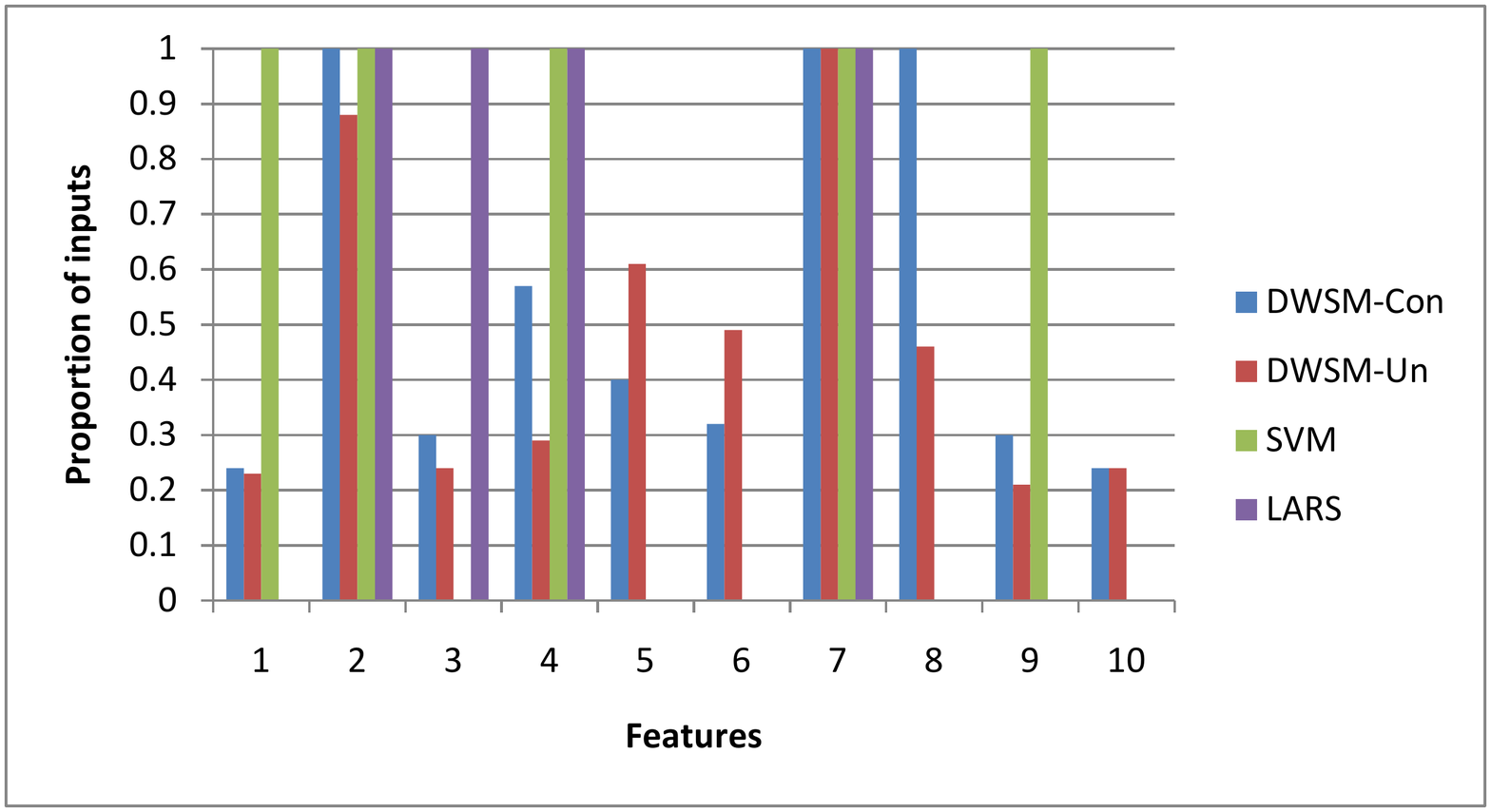} & \hspace{-1cm}  \includegraphics[width=0.6\linewidth]{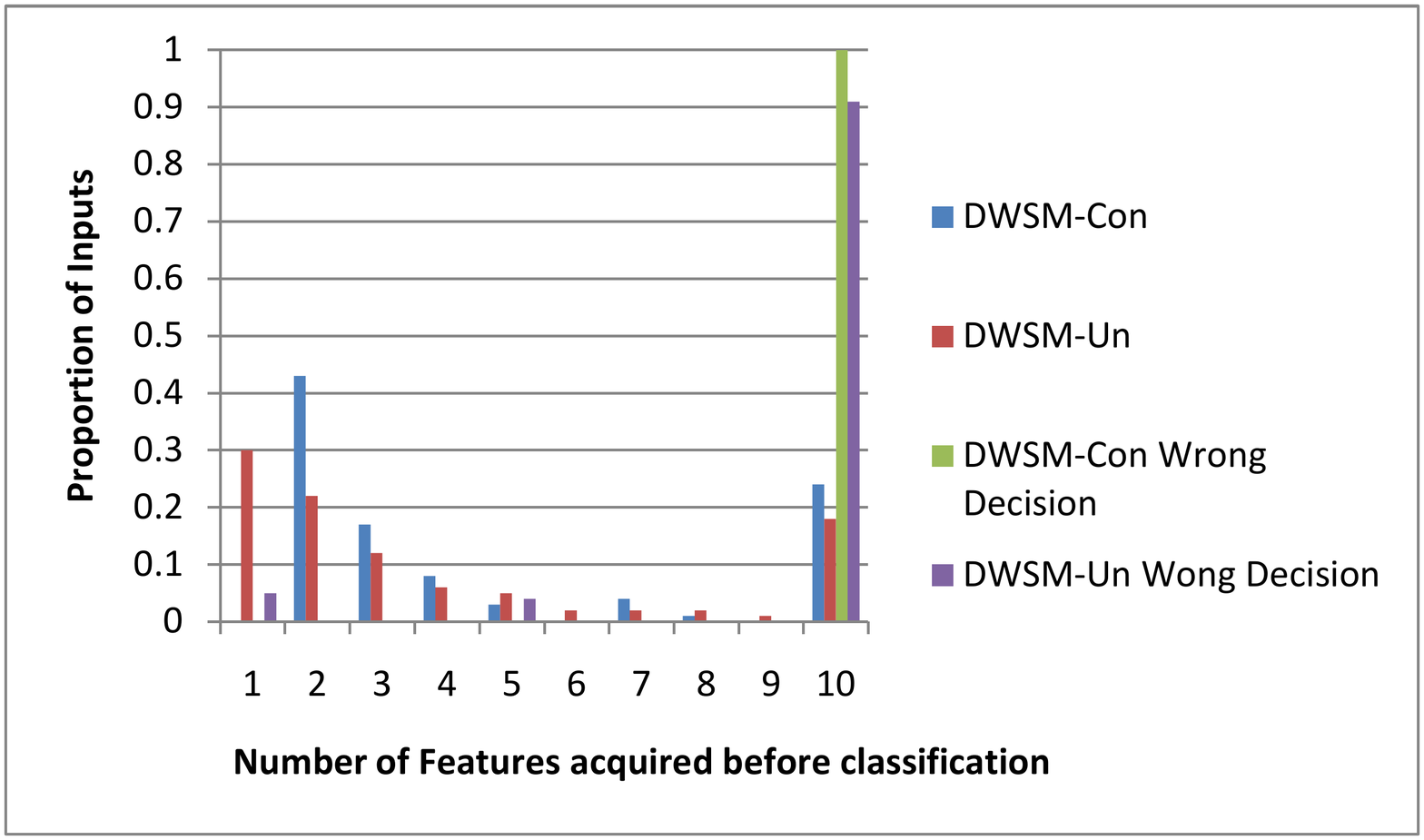}
\vspace{-0.5cm}
\end{tabular}
\end{center}
\vspace{-0.5cm}
\caption{\textit{Breast-Cancer, training size = 10\%, Sparsity $\approx$ 50 \%} \textbf{Left: } The distribution of use of each feature. For example, DWSM-Con uses feature $2$ for classifying 100\% of the test examples, while DWSM-Un uses this feature for classifying only 88\% of the examples. \textbf{Right: } The mean proportion of features used for classifying. For example DWSM-Con classifies 42\% of the examples using exactly 2 features while DWSM-Un classifies 21\% of the examples using exactly 2 features.}
\label{fig:dist}
\end{figure}
\vspace{-0.5cm}

Figure \ref{fig:dist} gives some qualitative results. First, from the left histogram, one can see that some features are used in 100\% of the decisions. This illustrates the ability of the model to detect important features that must be used for decision.  Note that many of these features are also used by the $L_1$-SVM and the LARS models. The sparsity gain in comparison to the baseline model is obtained through the features 1 and 9 that are only used in about 20\% of decisions. From the right histogram, one can see that the DWSM model mainly classifies using 1, 2, 3 or 10 features, showing that the model is able to adapt its behaviour to the difficulty of classifying a particular input. This is confirmed by the green and violet histograms that show that for incorrect decisions (i.e. very difficult inputs) the classifier almost always acquires all the features before classifying. These difficult inputs seem to have been identified, but the set of features is not sufficient for a good understanding. This behaviour opens appealing research directions concerning the acquisition and creation of new features (see Section \ref{part:conc}).

\section{Related Work}
\vspace{-0.35cm}
\label{sec:relatedWork}

\label{sec:sota}

Feature selection comes in three main flavors \cite{Guyon2003}: wrapper,
filter, or embedded approaches. \textbf{Wrapper approaches} involve searching the feature space for an optimal subset of features that maximize classifier performance.  The feature selection step wraps around the classifier, using the classifier as a black-box evaluator of the selected feature subset. Searching the entire feature space is very quickly intractable and therefore various approaches have been proposed to restrict the search (see \cite{eurogp2008,Gaudel2010}).  The advantage of the wrapper approaches is that the feature subset decision can take into consideration feature inter-dependencies and avoid redundant features, however the problem remains of the exponential size of the search space. \textbf{Filter approaches} rank the features by some scoring function independent of their effect on the associated classifier.  Since the choice of features is not influenced by classifier performance, filter approaches rely purely on the adequacy of their scoring functions. Filtering methods are susceptible to not discriminating redundant features, and missing feature inter-dependencies (since each feature is scored individually).  Filter approaches are however easier to compute and more statistically stable relative to changes in the dataset. \textbf{Embedded approaches} include feature selection as part of the learning machine. These include algorithms solving the LASSO problem \cite{Tibshirani1994}, and other linear models involving a regularizer based on a sparsity inducing norm ($\ell_{p \in   [0;1]}$-norms \cite{Xuetal2010}, group LASSO, ...). Kernel machines provide a mixture of feature selection and construction as part of the classification problem. Decision trees are also considered embedded approaches although they are also similar to filter approaches in their use of heuristic scores for tree construction. The main critique of embedded approaches is two-fold: they are susceptible to include redundant features, and not all the techniques described are easily applied to multi-class problems.In brief, both filtering and embedded approaches have their drawbacks in terms of their ability to select the best subset of features, whereas wrapper methods have their main drawback in the intractability
of searching the entire feature space. Furthermore, all existing methods perform feature selection based on the whole training set, the same set of features being used to represent any data.

Our sequential decision problem defines both feature selection and classification tasks. In this sense, our approach resembles an embedded approach. In practice, however, the final classifier for each single datapoint remains a separate entity, a sort of black-box classifying machine upon which performance is evaluated. Additionally, the learning algorithm is free to navigate over the entire combinatorial feature space. In this sense our approach resembles a wrapper method.

There has been some work using similar formalisms \cite{Ertin2002}, but with different goals and lacking in experimental results.  Sequential decision approaches have been used for cost-sensitive classification with similar models \cite{Ji2007}.  There have also been applications of Reinforcement Learning to optimize anytime classification \cite{Poczos2009}.  We have previously looked at using Reinforcement Learning for finding a stopping point in feature quantity during text classification \cite{Dulac-Arnold2011}.

Finally, in some sense, DWSC has some similarity with decision trees as each new datapoint that is labeled is following a different path in the feature space. However, the underlying mechanism is quite different both in term of inference procedure and learning criterion. There has been some  work in using RL for generating decision trees \cite{Preda2007}, but that approach is still tied to decision tree construction heuristics and the end product remains a decision tree.


\section{Conclusion}
\vspace{-0.35cm}
\label{part:conc}
In this article we introduced the concept of datum-wise classification, where we learn both a classifier and a sparse representation of the data that is adaptive to each new datum being classified.  We took an approach to sparsity that considers the combinatorial space of features, and proposed a sequential algorithm inspired by Reinforcement Learning to solve this problem.  We showed that finding an optimal policy for our Reinforcement Learning problem is equivalent to minimizing the $L_0$ regularized loss of our classification problem.  Additionally we showed that our model works naturally on multi-class problems, and is easily extended to avoid overfitting on datasets where the number of features is larger than the number of examples. Experimental results on 14 datasets showed that our approach is indeed able to increase sparsity while maintaining equivalent classification accuracy.




%


%
\vspace{-0.4cm}
\bibliographystyle{IEEEtran}
\bibliography{bibAux}
\end{document}